\documentclass[10pt,twocolumn,letterpaper]{article}

\usepackage[pagenumbers]{wacv} % To force page numbers, e.g. for an arXiv version

\usepackage{amsmath}
\usepackage{algorithm}
\usepackage{algpseudocode}
\usepackage{lipsum}
\usepackage{multirow, multicol}
\usepackage{array}
\usepackage{graphicx}
\usepackage{wrapfig}
\newcolumntype{L}[1]{>{\raggedright\let\newline\\\arraybackslash\hspace{0pt}}m{#1}}
\newcolumntype{C}[1]{>{\centering\let\newline\\\arraybackslash\hspace{0pt}}m{#1}}
\newcolumntype{R}[1]{>{\raggedleft\let\newline\\\arraybackslash\hspace{0pt}}m{#1}}
\usepackage{caption}
\usepackage{subcaption}
\usepackage{amssymb}% http://ctan.org/pkg/amssymb
\usepackage{pifont}% http://ctan.org/pkg/pifont
\newcommand{\cmark}{\ding{51}}%
\newcommand{\mycomment}[1]{}

\algrenewcommand\algorithmicindent{1em}%

\algrenewcommand\algorithmicrequire{\textbf{Input:}}
\algrenewcommand\algorithmicensure{\textbf{Output:}}
\usepackage{float}
\usepackage{colortbl}
\usepackage{caption}
\definecolor{wacvblue}{rgb}{0.21,0.49,0.74}
\usepackage[pagebackref,breaklinks,colorlinks,allcolors=wacvblue]{hyperref}

\def\wacvPaperID{***} % *** Enter the WACV Paper ID here
\def\confName{WACV}
\def\confYear{2026}

\title{Test-Time Adaptation for Video Highlight Detection Using Meta-Auxiliary Learning and Cross-Modality Hallucinations}

\author{Zahidul Islam$^{1}$ \qquad\qquad Sujoy Paul$^{2}$ \qquad\qquad Mrigank Rochan$^{1}$ \\
	$^1$University of Saskatchewan, Canada \qquad $^2$Google DeepMind
}

\begin{document}
\maketitle
\begin{abstract}
    Existing video highlight detection methods, although advanced, struggle to generalize well to all test videos. These methods typically employ a generic highlight detection model for each test video, which is suboptimal as it fails to account for the unique characteristics and variations of individual test videos. Such fixed models do not adapt to the diverse content, styles, or audio and visual qualities present in new, unseen test videos, leading to reduced highlight detection performance. In this paper, we propose Highlight-TTA, a test-time adaptation framework for video highlight detection that addresses this limitation by dynamically adapting the model during testing to better align with the specific characteristics of each test video, thereby improving generalization and highlight detection performance. Highlight-TTA is jointly optimized with an auxiliary task, cross-modality hallucinations, alongside the primary highlight detection task. We utilize a meta-auxiliary training scheme to enable effective adaptation through the auxiliary task while enhancing the primary task. During testing, we adapt the trained model using the auxiliary task on the test video to further enhance its highlight detection performance. Extensive experiments with three state-of-the-art highlight detection models and three benchmark datasets show that the introduction of Highlight-TTA to these models improves their performance, yielding superior results.

\end{abstract}
\section{Introduction}
Video highlight detection involves identifying and extracting the most significant and engaging moments from video content \cite{badamdorj2021joint, badamdorj2022contrastive, hong2020mini, xiong2019lessLM, moon2023qddetr, sun2024trdetr}. This technology is essential in various fields such as sports, entertainment, education, and social media, where it enhances the user experience by providing quick access to the most relevant parts of a video. By automatically pinpointing key events, video highlight detection saves viewers time and ensures they can easily find and enjoy the most compelling segments. This capability is increasingly important in today's digital age, where vast amounts of video content are produced and consumed daily. However, in the rapidly evolving landscape of video content, accurately detecting highlights is becoming increasingly challenging. Videos can vary significantly in terms of content, context, and visual/audio quality, making it challenging for existing fixed and generic highlight detection models \cite{sun2014rankingdomainspecific,xiong2019lessLM,wang2020learningtrailer,badamdorj2021joint,badamdorj2022contrastive,liu2022umt,li2024unsupervised} to accurately detect highlights across all instances. To tackle this issue, test-time adaptation (TTA) \cite{sun2020test} provides a compelling solution. TTA is essential for video highlight detection as it enables the model to adapt to the specific characteristics of each video instance during inference. By adapting the model at test time, it can better account for these variations, leading to more accurate and robust highlight detection results tailored to each individual video. Therefore, we introduce Highlight-TTA, a test-time adaptation framework designed to advance video highlight detection, ensuring it can effectively handle diverse and ever-changing video content. To our knowledge, we are the first to explore test-time adaptation (TTA) in video highlight detection.

Auxiliary tasks are often used for test-time adaptation (TTA) to provide additional supervision and improve the generalization of the model \cite{sun2020test,chi2021test,hatem2023point,hatem2023test}. By leveraging related but easier-to-solve tasks, the model can learn more robust and transferable representations, which can help improve performance on the primary task, especially when test data differs from training data. In this work, we present a new self-supervised auxiliary task, called cross-modality hallucinations, for video highlight detection. Visual and audio modalities provide complementary information, and recent studies have shown that combining these modalities leads to superior highlight detection performance compared to using a single modality \cite{badamdorj2021joint,liu2022umt}. Our Highlight-TTA framework utilizes the available visual and audio modalities within a video to learn to hallucinate the features of one modality from the other. By performing cross-modality hallucinations, the model can infer missing or obscured information in one modality based on cues from the other, leading to a deeper understanding of the correlation between audio and visual components and thereby enhancing its performance on the primary task of highlight detection.

While we could jointly train a model using both the auxiliary and primary tasks, recent studies \cite{chi2021test,hatem2023point} and our experiments suggest that this may not be optimal, as it can be biased towards improving the auxiliary task at the expense of the primary task. To address this, we employ a meta-auxiliary learning approach based on model-agnostic meta-learning (MAML) \cite{finn2017model} to enable effective test-time adaptation for highlight detection. During training, we have access to the ground-truth highlights, and each video-highlight pair is treated as a task in our meta-learning approach. For each video, we first update the model using the auxiliary loss derived from cross-modality hallucinations. This updated model is then used to perform the primary highlight detection task, with the highlight detection loss subsequently used to update the model. The model parameters are optimized such that the updates from the auxiliary task enhance the performance of the primary highlight detection task. Finally, at test time, our model can effectively adapt to each test video when fine-tuned using the auxiliary task, ensuring improved highlight detection performance.

In summary, this paper makes the following contributions: 1) We propose Highlight-TTA, a test-time adaptation framework for video highlight detection. To the best of our knowledge, this is the first work to apply test-time adaptation for video highlight detection; 2) We introduce a new self-supervised auxiliary task, cross-modality hallucinations, to improve the performance of the primary highlight detection task; 3) We propose a meta-auxiliary learning scheme to optimize the model parameters, ensuring that adapting these parameters through the auxiliary task during testing improves highlight detection performance; and 
4) We implement our Highlight-TTA framework on top of three state-of-the-art highlight detection models and demonstrate its effectiveness through extensive experiments on three benchmark datasets.

\section{Related Work}
Driven by the growing demand for efficient content viewing and summarization, many approaches have been proposed to tackle video highlight detection, spanning from traditional rule-based techniques to cutting-edge deep learning based methods. Early works often relied on handcrafted features and heuristics to identify key moments or segments within videos \cite{song2015tvsum}. However, with the advent of deep learning, researchers have shifted towards data-driven approaches for highlight detection that learn representations directly from raw video data. Many of high performing highlight detection approaches rely on manually annotated frame-level supervision \cite{gygli2016video2gif, jiao2018deepranking, wang2020learningtrailer, yu2018deep, rochan2020adaptive, badamdorj2021joint, liu2022umt, moon2023qddetr, yang2024taskweave, lin2023univtg, sun2024trdetr, xiao2024bridging_uvcom}. To avoid costly labeled data, some methods leverage cheaper video-level tags or category  for weak supervision \cite{yang2015unsupervisedRRAE, cai2018weaklyVESD, hong2020mini, panda2017weaklyDSN,xiong2019lessLM, ye2021temporal}. However, these approaches often require access to large external datasets for training. Notably, some recent unsupervised methods have also shown promising results  \cite{badamdorj2022contrastive,li2024unsupervised, 
islam2024unsupervisedvideohighlightdetection}. However, existing methods typically struggle to generalize well since they focus on a generic highlight detection model, which suffers when applied directly to new, unseen testing videos due to distribution shifts between training and testing data. To address this, inspired by recent success of test-time adaptation \cite{sun2020test}, we propose Highlight-TTA, a test-time adaptation framework for video highlight detection.

Our work is related to a popular meta-learning algorithm, MAML \cite{finn2017model}, which enables rapid learning and adaptation to new tasks using only a few training examples. Additionally, our work is connected to the meta-auxiliary learning framework (MAXL) \cite{liu2019self}, which generates auxiliary labels to enhance the primary task using ground-truth labels. Recent studies in point clouds \cite{hatem2023point, hatem2023test} and image deblurring \cite{chi2021test} have explored meta-auxiliary learning combined with test-time adaptation. To our knowledge, we are the first to explore test-time adaptation using meta-auxiliary learning for video highlight detection. Additionally, we introduce a new self-supervised auxiliary task, cross-modal hallucinations, to leverage multimodal audio-visual information for effective video highlight detection.

\section{Our Approach}\label{sec:app}

We present Highlight-TTA, a test-time adaptation framework for highlight detection that leverages audio and visual modalities to adapt and enhance the highlights of a video.

Given a video $V$, we split it into $n$ clips, with each clip containing a fixed number of frames. From each clip $c_i$ (where $i=1,2,...n$), we extract its corresponding visual features $v_i \in \mathbb{R}^{d_v}$ from a pre-trained visual feature extractor and audio features $a_i \in \mathbb{R}^{d_a}$ from a pre-trained audio feature extractor. We denote $V$ with its visual features, $\{v_i\}_{i=1}^n$, and audio features, $\{a_i\}_{i=1}^n$. Our Highlight-TTA method aims to find a model $F_\theta(V)\rightarrow H$, parameterized by $\theta$ that maps $V$ to a set of highlight scores $H = \{h_i\}_{i=1}^n$, where $h_i$ indicates the highlight score of each clip $c_i$ in video $V$. 
Our method utilizes a meta-auxiliary learning mechanism together with a self-supervised auxiliary task called cross-modality hallucinations that enable effective and fast adaptation to each test instance.

\subsection{Network Architecture}\label{subsec:net}
We implement Highlight-TTA on top of three  three state-of-the-art multimodal highlight detection models, JAV \cite{badamdorj2021joint}, UMT \cite{liu2022umt}, and QD-DETR \cite{moon2023qddetr}, which are based on Transformers \cite{attentionvaswani}, and at their core, utilize unimodal modules to encode temporal dependencies within the same modality and bimodal modules to capture cross-modal relationships. We extend these networks by introducing two hallucination modules with learnable parameters next to the unimodal modules for our self-supervised auxiliary task. We denote the parameters of our model $F_\theta(V)$ as $\theta = \{\theta^{s}, \theta^{p}, \theta^{a}\}$. $\theta^{p}$ are the parameters involved only in the primary task and $\theta^{a}$ represent the parameters of two cross-modality hallucination modules (see Sec. \ref{subsec:cmh}), which take part only in the auxiliary task. The parameters involved in both the primary and auxiliary tasks are denoted as $\theta^{s}$.

\begin{figure}[!h]
    \centering
    \includegraphics[width=0.375\textwidth]{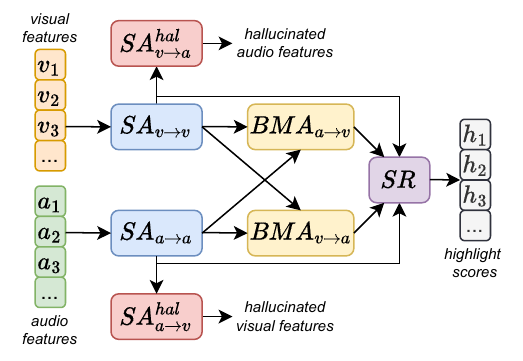} 
    \caption{A schematic overview of our model built upon JAV \cite{badamdorj2021joint} for the proposed Highlight-TTA framework.}
    \label{fig:network}
\end{figure}

Fig. \ref{fig:network} illustrates a schematic overview of our model built upon JAV \cite{badamdorj2021joint}. 
As illustrated in Fig \ref{fig:network}, given a video with $n$ clips, our audio-visual model first processes the clip-level visual features $\{v_i\}_{i=1}^n$ using a self-attention layer $\text{SA}_{v \rightarrow v}$ and the clip-level audio features $\{a_i\}_{i=1}^n$ using another self-attention layer $\text{SA}_{a \rightarrow a}$. Then, we feed the self-attended visual features $\{v_i^v\}_{i=1}^n$ into our cross-modal audio hallucination module $\text{SA}^{hal}_{v \rightarrow a}$ to hallucinate self-attended audio features $\{a_i^a\}_{i=1}^n$. Similarly, we send the self-attended audio features  $\{a_i^a\}_{i=1}^n$ to our cross-modality visual hallucination module $\text{SA}^{hal}_{a \rightarrow v}$, to hallucinate the self-attended visual features $\{v_i^v\}_{i=1}^n$. 
Each of these hallucination modules comprises a self-attention layer sandwiched between two fully-connected layers and includes a skip connection that bypasses the self-attention layer. 
On the other hand, two bimodal attention layers $\text{BMA}_{v \rightarrow a}$ and $\text{BMA}_{a \rightarrow v}$ are also fed the self-attended visual features $\{v_i^v\}_{i=1}^n$ and self-attended audio features $\{a_i^a\}_{i=1}^n$ to produce bimodal attended features, $\{v_i^a\}_{i=1}^n$ and $\{a_i^v\}_{i=1}^n$, respectively. Finally, a score regressor module (SR) combines the self-attended and bimodal attended features using a set of learnable weights and passes them through two fully-connected layers to predict the highlight score ${h_i}$ for each clip in the video $V$. 

In addition to JAV \cite{badamdorj2021joint}, we apply Highlight-TTA to two other state-of-the-art highlight detection models: UMT \cite{liu2022umt} and QD-DETR \cite{moon2023qddetr}. UMT is a multimodal, transformer-based model with encoders for audio, visual, and text modalities. The features from these unimodal encoders are fused using a cross-modal encoder to capture cross-modal dependencies. Since, for Highlight-TTA, we focus on conventional highlight detection that solely relies on vidoes and not on user preferences through text queries, we utilize only audio and visual modalities. Based on this, we extend UMT by adding two cross-modal hallucination modules, similar to JAV, while excluding the text modality. QD-DETR is primarily a text-visual model with one stream for processing text features and another for encoding visual features. It uses a cross-attention transformer to fuse text and visual features. We replace its text input features with audio features and introduce cross-modal hallucination modules.

%--------------------

\subsection{Cross-Modality Hallucinations}\label{subsec:cmh}
We introduce cross-modality hallucinations, a self-supervised auxiliary task in our Highlight-TTA framework. Videos often encode rich information in both audio and visual modalities, and their fusion can provide valuable insights into the underlying events and highlights \cite{badamdorj2021joint}. By training a model to hallucinate one modality from the other (i.e., from visual to audio and vice versa), we facilitate a deeper understanding of the inherent correlations between audio and visual components. For instance, a surge in crowd noise often corresponds to exciting visual moments in sports videos. Through cross-modality hallucinations, the model can infer missing or obscured information in one modality based on cues from the other, thereby enhancing its ability to identify meaningful highlights. Furthermore, incorporating cross-modality hallucinations addresses the challenge of distribution shifts at test time, as the model becomes adept at extracting and integrating multi-modal information, leading to more accurate highlight detection.

In cross-modal hallucination, given features from one modality, the goal is to approximate or hallucinate features from the other modality. From the two cross-modal hallucination modules in our network (see Sec. \ref{subsec:net}), we compute the cross-modal visual hallucination loss, \(L^{hal}_{a \rightarrow v}\), and the cross-modal audio hallucination loss, \(L^{hal}_{v \rightarrow a}\). 
% As shown in Eq. \ref{eq:hal_av}, 
\(L^{hal}_{a \rightarrow v}\) is calculated using a Mean Squared Error (MSE) loss between the output hallucinated features of our cross-modal visual hallucination module, \(\text{SA}^{hal}_{a \rightarrow v}\), and the self-attended visual features, \(\{v_i^v\}_{i=1}^n\). Note that we detach the gradients of \(\{v_i^v\}_{i=1}^n\) before using them to calculate the cross-modal visual hallucination loss, ensuring that backpropagation occurs only through the layers in the audio-modality branch. 
Similarly, we calculate the cross-modal audio hallucination loss, \(L^{hal}_{v \rightarrow a}\). We add these two cross-modal hallucination losses to compute our auxiliary loss, \(L_{aux}\).

\subsection{Joint Training}\label{subsec:jt}
We calculate the primary loss of our network, $L_{pri}$, as a binary cross-entropy loss between the predicted highlight scores, $H$, and the ground-truth highlight scores, $H^{gt}$. We first train our network using both our primary and auxiliary losses jointly. As shown in Eq. \ref{eq:joint}, the loss, $L_{joint}$, for this joint training of our model can be formulated as the combination of $L_{pri}$, which requires the ground-truth highlight annotation $H^{gt}$ along with the input video $V$ and updates both the primary parameters, $\theta^{p}$, and shared parameters, $\theta^{s}$, and the auxiliary loss, $L_{aux}$, which only requires the input video $V$ and updates the parameters $\theta^{s}$ and $\theta^{a}$.

{\small{
\begin{equation}
\label{eq:joint}
L_{joint} = L_{pri}(\{\theta^{s}, \theta^{p}\}; V, H^{gt}) + L_{aux}(\{\theta^{s}, \theta^{a}\}; V)
\end{equation}}}

We utilize the output model of this joint-training step as an initialization for the meta-auxiliary training stage.

\begin{figure*}[t]
    \centering \includegraphics[width=0.9\textwidth]{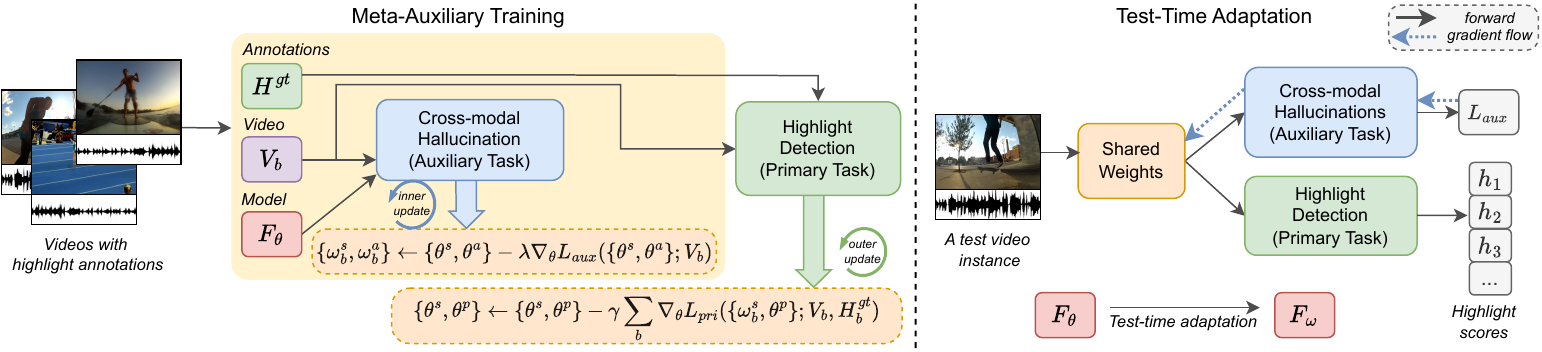}
    \caption{Illustration of our meta-auxiliary training and test-time adaptation method for video highlight detection. During the meta-auxiliary training stage, we initially obtain adapted parameters using the auxiliary loss present in the inner loop. Next, we evaluate and update the model weights in the outer loop by using the primary loss computed from the adapted parameters. At test-time, we update the model using the auxiliary task to adapt specifically to a given test video instance.}
    \label{fig:meta}
\end{figure*}

\subsection{Meta-Auxiliary Training}\label{subsec:meta}
The joint-trained model from Eq. \ref{eq:joint} can be directly used for testing, but it is sub-optimal as it does not adapt to the internal information in test video instances. While we could use the auxiliary task to fine-tune the joint-trained model at test-time, there is a risk that the model becomes more biased towards improving the auxiliary task \cite{chi2021test,hatem2023point,hatem2023test,sun2020test} rather than the primary task. Therefore, following prior work \cite{chi2021test,hatem2023point,hatem2023test}, we adopt a meta-auxiliary training strategy which involves updating the model using both auxiliary and primary tasks in a manner that aids test-time adaptation to specific test instance using the auxiliary task. Specifically, for each test video, our model is updated and adapted accordingly.

%-------------------------
Our meta-auxiliary training algorithm (Alg. \ref{alg:training} and Fig. \ref{fig:meta}) has two loops. In the inner loop, given a video \(V_b\) and its ground-truth \(H^{gt}_b\) from a batch of \(B\) videos, along with our joint-trained model parameters \(\theta\) (which we obtain by training the primary and auxiliary tasks together as described in Sec. \ref{subsec:jt}), we first make a few gradient updates using the auxiliary loss, \(L_{aux}(\{\theta^{{s}}, \theta^{a} \}; V_b)\). During these inner loop updates, we obtain the updated model parameters \(\{\omega^{s}_b, \omega^{a}_b\}\) shown in Eq. \ref{eq:auxiliary_update}, where \(\lambda\) is the inner learning rate. This inner loop adaptation can be performed during test time as it does not rely on ground-truth highlights.

{\small{
\begin{equation}
  \label{eq:auxiliary_update}
    \{\omega^{s}_b, \omega^{a}_b\} \leftarrow \{\theta^{{s}}, \theta^{a} \} - \lambda \nabla_{\theta} L_{aux}
\end{equation}
}}

The adapted parameters \(\omega^{s}_b\) from the inner updates can then be used for our primary task of highlight detection. This step ensures that the primary task is highly coupled with the updates in the auxiliary task. This
encourages our model to be updated in a way that, once adapted to a given video, it enhances the performance of the primary task. As shown in Eq. \ref{eq:primary_update}, in outer update of the algorithm, we optimize \(\theta\) based on the primary loss, \(L_{pri}(\{\omega^{s}_b, \theta^{p}\}; V_b, H^{gt}_b)\). Here, \(\gamma\) is the meta-learning rate.

{\small{
\begin{equation}
\label{eq:primary_update}
\{\theta^{s}, \theta^{p}\} \leftarrow \{\theta^{s}, \theta^{p}\} - \gamma \sum_{b=1}^{B} \nabla_{\theta} L_{pri}
\end{equation}
}}

As the primary branch involves $\theta^{s}$ and $\theta^{p}$, only these parameters are updated using \(L_{pri}\) in the outer loop. Whereas, $\theta^{a}$ is updated using the auxiliary loss in the inner loop.

\begin{algorithm}
% \small
\footnotesize
% \scriptsize
\caption{\small{Our Meta-Auxiliary Training Algorithm}}
\label{alg:training}
\begin{algorithmic}[1]
\Require $\lambda$: inner learning rate, $\gamma$: meta learning rate, 

$V$: video, $H^{gt}$: ground-truth highlights
\Ensure $\theta$: learned parameters from meta-auxiliary training
\State Initialize the model with joint-trained weights: 

$\theta = \{\theta^{s}, \theta^{{a}}, \theta^{{p}}\}$

\While{not converged}
    \State Sample a batch of training instances $\{V_b, H^{gt}_b\}_{b=1}^B$
    \For{each $b$}
        \State Compute auxiliary cross-modality hallucinations loss: 
        
        $L_{{aux}} = L^{hal}_{a \rightarrow v} + L^{hal}_{v \rightarrow a}$

        \State Compute adapted parameters: 
        
        $\{\omega^{s}_b, \omega^{a}_b\} \leftarrow \{\theta^{{s}}, \theta^{a}\} - \lambda \nabla_{\theta} L_{aux}(\{\theta^{{s}}, \theta^{a}\}; V_b)$

        \State Auxiliary task update: 
        
        $\{\theta^{{s}}, \theta^{a}\} \leftarrow \{\theta^{{s}}, \theta^{a}\} - \lambda \nabla_{\theta} L_{aux}(\{\theta^{{s}}, \theta^{a}\}; V_b)$

    \EndFor
    \State Primary task update: 
    
    \noindent\hspace{8pt}$\{\theta^{s}, \theta^{p}\} \leftarrow \{\theta^{s}, \theta^{p}\} - \gamma \sum_{b}^{} \nabla_{\theta} L_{pri}(\{\omega^{s}_b, \theta^{p}\}; V_b, H^{gt}_b)$
\EndWhile
\end{algorithmic}
\end{algorithm}

\subsection{Test-Time Adaptation}\label{subsec:tta}
At test-time, we utilize the model obtained from meta-auxiliary training with parameters \(\theta\) to initialize test-time training. Given a test video instance \(V_b\), we evaluate the auxiliary loss \(L_{aux}\), which we use to adapt (using Eq. \ref{eq:auxiliary_update}) the model to \(V_b\) and obtain adapted parameters \(\omega^{s}_b\). We use these adapted parameters \(\omega^{s}_b\), along with the parameters \(\theta^{p}\), to detect highlights of video \(V_b\).

\section{Experiments}\label{sec:exp}
\subsection{Datasets and Settings} \label{subsec:exp_setup}
{\noindent\textbf{Datasets:}} We utilize three benchmark video highlight detection datasets, namely YouTube \cite{sun2014rankingdomainspecific}, TVSum \cite{song2015tvsum}, and QVHighlights \cite{lei_moment_detr}. YouTube includes videos from six categories with approximately 100 videos per category. We follow the standard train-test splits provided with the dataset. TVSum is a smaller dataset with 50 videos across 10 categories. We follow prior works \cite{badamdorj2021joint, badamdorj2022contrastive, rochan2019videounpaired} and utilize a random train-test split with a ratio of 80:20. We run our experiments on TVSum five times and report the average performance. QVHighlights, a large dataset containing about 10,000 videos, is primarily designed for query-focused video highlight detection and moment retrieval. Each video has corresponding textual queries and saliency/highlight scores. The dataset has canonical train, validation, and test splits with a ratio of 70:15:15. Since our method only requires videos, we ignore the user query annotations. For a fair comparison, we evaluate our method against prior non-query-based methods on this dataset.

{\noindent\textbf{Features:}} Following prior work on TVSum and YouTube \cite{badamdorj2022contrastive, badamdorj2021joint}, we extract visual features from each clip using a 3D-CNN based on ResNet-34 architecture. For QVHighlights videos, following \cite{lei_moment_detr, liu2022umt}, we extract visual features using SlowFast \cite{feichtenhofer2019slowfast} and the CLIP video encoder (ViT-B/32) \cite{clip_radford2021learning}. To extract audio features, on all datasets, we use PANN \cite{kong2020panns} audio network pre-trained on AudioSet \cite{audioset}.

\noindent{\textbf{Backbones:}} We implement Highlight-TTA on top of three state-of-the-art highlight detection methods, namely JAV \cite{badamdorj2021joint}, UMT \cite{liu2022umt}, and QD-DETR \cite{moon2023qddetr}, for which we could obtain the code and reproduce their experiments. To apply Highlight-TTA on these methods, we introduce cross-modal hallucination modules, which are utilized for meta-auxiliary training and TTA. Note that we modified UMT, which originally uses visual, audio, and textual information, to exclude textual information. For QD-DETR, we replace the textual input features with audio features, to obtain its audio-visual variant, making it compatible for Highlight-TTA.

% Main results TVSum
\begin{table*}[!h]
\setlength{\tabcolsep}{4pt}
\centering
\footnotesize
    \begin{tabular}{ccccccc|c >{\columncolor[gray]{0.9}}c|c >{\columncolor[gray]{0.9}}c|c >{\columncolor[gray]{0.9}}c}
    \hline
        VESD  & LM &  Trail. & CHD & CHD &  MT & UAVR & JAV & \textbf{JAV + Ours} & UMT & \textbf{UMT + Ours} & QD-DETR & \textbf{QD-DETR + Ours}\\
        (V) & (V) & (V) & (V) & (AV) & (AV) & (AV)  & (AV) & (AV) & (AV) & (AV) & (AV) & (AV) 
    \\ \hline
    48.10 & 56.40 & 62.80  & 52.76 & 55.15 & 78.30 & 60.34 & 68.42 & \textbf{71.05} & 80.48 & \textbf{80.96} & 63.66 & \textbf{67.81}\\ \hline
    
    \end{tabular}
    \caption{Highlight detection results (top-5 mAP) on TVSum.}
    \label{table:comparison_tvsum}
\end{table*}

% Main YOUTUBE
\begin{table*}[!h]
\setlength{\tabcolsep}{4pt}
\centering
\footnotesize
    \begin{tabular}{ccccccc|c >{\columncolor[gray]{0.9}}c|c >{\columncolor[gray]{0.9}}c|c >{\columncolor[gray]{0.9}}c}
    \hline
      LSVM & LM  &  Trail. & CHD & CHD & MT & UAVR & JAV  & \textbf{JAV + Ours}& UMT & \textbf{UMT + Ours} & QD-DETR & \textbf{QD-DETR + Ours}\\
    (V) & (V) & (V) & (V) & (AV) & (AV) & (AV) & (AV) & (AV) & (AV) & (AV) & (AV) & (AV)    
    \\ \hline
   53.60 & 56.40 & 69.10 & 65.39 & 65.72 & 65.10 & 68.30 & 70.18 & \textbf{73.02} & 74.86 & \textbf{75.63} & 68.19 & \textbf{70.04}\\ \hline
   
    \end{tabular}
    \caption{Highlight detection results (mAP) on YouTube.}
    \label{table:comparison_youtube}
\end{table*}

% Main QVHIGHLIGHTS
\begin{table*}[!h]
\setlength{\tabcolsep}{5pt}
\centering
\footnotesize
    \begin{tabular}{lcccc|c >{\columncolor[gray]{0.9}}c|c >{\columncolor[gray]{0.9}}c|c >{\columncolor[gray]{0.9}}c}
    \hline
    \multirow{2}{*}{Metric}  & BT  & CHD & CHD & UAVR &JAV & \textbf{JAV + Ours} & UMT & \textbf{UMT + Ours}  & QD-DETR & \textbf{QD-DETR + Ours}\\
    & (V) & (V) & (AV) & (AV) & (AV) & (AV) & (AV) & (AV) & (AV) & (AV) \\ \hline
    mAP     & 14.36 & 15.82 & 17.25 & 18.38 & 23.98 & \textbf{24.74} & 24.37 & \textbf{24.51} & 24.23	 & \textbf{24.64}\\
    HIT@1     & 20.88 & 17.10 & 18.60 & 24.71 & 29.70 & \textbf{31.19} & 31.13 & \textbf{31.39} & 30.09	 & \textbf{31.06}\\ \hline
    \end{tabular}
    \caption{Results (mAP and HIT@1) on QVHighlights \textit{test} set. We report results obtained from their evaluation server.}
    \label{table:comparison_qvhighlights}
\end{table*}

\noindent{\textbf{Evaluation Metrics:}} On QVHighlights, we use mean average precision (mAP) for evaluation, which takes into account the highlight scores of all clips, and HIT@1, which considers the hit ratio of the clip with the highest score for each video. Following prior work \cite{lei_moment_detr}, we consider only the clips rated as \textit{Very Good} by users to be highlights. Following existing work \cite{badamdorj2022contrastive, badamdorj2021joint}, on YouTube, we report mAP, and on TVSum, we report mAP on the top five predicted clips (top-5 mAP). We report all metrics as percentages. 

\noindent{\textbf{Implementation Details:}} We build upon the official code of JAV, UMT, and QD-DETR for Highlight-TTA. For both JAV and UMT, in the outer loop update of meta-auxiliary training (Alg. \ref{alg:training}), we use the Adam optimizer \cite{kingma2014adam} with a learning rate (LR) of $\gamma = 5\times10^{-5}$. We use the same optimizer and learning rate for joint-training as well. For the inner update of our meta-auxiliary training, we use the SGD optimizer with a LR of $\lambda = 1\times10^{-1}$. During test-time adaptation with UMT, we use LR of $\lambda = 1\times10^{-1}$ to update our model. For JAV, during test-time adaptation, we use LR of $\lambda = 1\times10^{-1}$ for the QVHighlights dataset and a LR of $\lambda = 5.5\times10^{-2}$ for TVSum  and YouTube. For QD-DETR, we use $\gamma = 1\times10^{-4}$ as the LR for the outer loop of meta-auxiliary training. $\lambda = 5\times10^{-2}$ is used as the LR for the inner update and test-time adaptation. 
We employ three gradient updates during training and testing to adapt our model using the auxiliary loss in Line 6 of Algo.~\ref{alg:training}.
For the JAV model, on YouTube, we train for 30 epochs during joint training and 10 epochs during meta-auxiliary learning. On the TVSum dataset, we train for 100 epochs during joint training and 20 epochs for meta-auxiliary learning. On QVHighlights, we train for 15 epochs in both phases. For UMT, we train for 100 epochs on the YouTube dataset during both joint training and meta-auxiliary learning. On the TVSum dataset, we train for 200 epochs in both phases, and on QVHighlights, we train for 50 epochs during both phases.
For QD-DETR, we train for 40 epochs during joint-training and 10 epochs for meta-auxiliary training on both YouTube and TVSum. On QVHighlights, we train for 10 epochs during joint-training and 5 epochs during meta-auxiliary training. We train our models on one NVIDIA GeForce RTX 2080 Ti 12GB GPU.

\noindent{\textbf{Comparison Mothods:}} We compare with state-of-the-art methods such as VESD \cite{cai2018weaklyVESD}, LSVM \cite{sun2014rankingdomainspecific}, LM \cite{xiong2019lessLM}, Trail. \cite{wang2020learningtrailer}, CHD \cite{badamdorj2022contrastive}, JAV \cite{badamdorj2021joint}, UMT \cite{liu2022umt},  QD-DETR \cite{moon2023qddetr}, MT \cite{li2024highlight}, and UAVR \cite{islam2024unsupervisedvideohighlightdetection} on TVSum and YouTube. 
For a fair comparison, on QVHighlights, we compare with prior methods that do not use textual queries: BT \cite{song2016beautythumb}, CHD \cite{badamdorj2022contrastive}, UAVR \cite{islam2024unsupervisedvideohighlightdetection}, JAV \cite{badamdorj2021joint}, UMT \cite{liu2022umt}, and QD-DETR \cite{moon2023qddetr}. Note that QD-DETR, primarily a text-video model, is adapted by replacing the text input with audio input. Additionally, for a stronger comparison, we extend CHD, a single-modal method that uses only visual features, by incorporating audio features using early fusion.

\subsection{Experimental Results}

{\flushleft \textbf{TVSum:}} In Table \ref{table:comparison_tvsum}, we report the highlight detection performance of Highlight-TTA on the TVSum dataset. Note that the results for CHD, JAV, UMT, and QD-DETR are from our own implementation and runs. The performance of all three state-of-the-art highlight detection models improves with Highlight-TTA, delivering superior results.

\noindent{\textbf{YouTube:}} Next, in Table \ref{table:comparison_youtube}, we report the highlight detection performance of Highlight-TTA on the YouTube dataset. Again, the performance of state-of-the-art highlight detection models improves with the introduction of our Highlight-TTA framework, yielding superior results.

\noindent{\textbf{QVHighlights:}} Finally, we report the performance of Highlight-TTA on the QVHighlights dataset in Table \ref{table:comparison_qvhighlights}. The results are based on the \textit{test} split, which must be accessed through their online evaluation server. For a fair comparison, we compare our approach with prior works that do not rely on query information, as discussed in Sec. \ref{subsec:exp_setup}. BT selects highlights based on the aesthetic quality and relevance of the video's visual content. CHD and JAV do not evaluate on QVHighlights, so we implemented these methods with their original settings and evaluated them on this dataset for comparison. As with the other datasets, the introduction of our Highlight-TTA framework improves the highlight detection performance of state-of-the-art methods.

\begin{figure*}[!h] 
\centering
\includegraphics[width=0.7\textwidth]{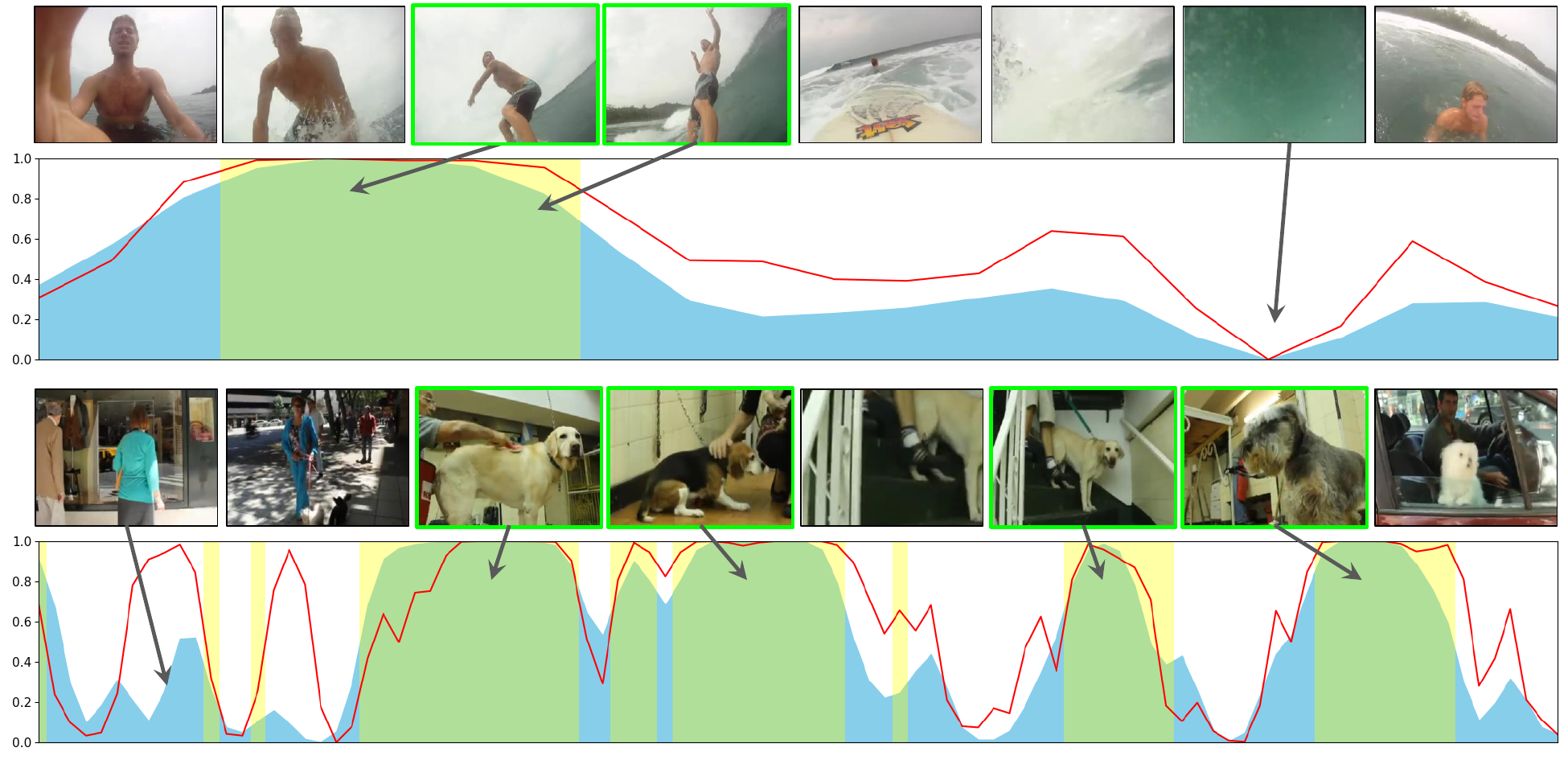} 
\caption{Visualization of highlight detection predictions using Highlight-TTA built upon JAV \cite{badamdorj2021joint} on an example test video from YouTube (top) and TVSum (bottom). The yellow region corresponds to ground-truth highlight annotations. We indicate the predicted scores of the joint-trained model with a red line, while the sky-blue region represents the predictions of our Highlight-TTA. With Highlight-TTA, the alignment of highlight predictions with ground-truth highlights improves, demonstrating the effectiveness of TTA for highlight detection.}
\label{fig:qual}
\end{figure*}

In Fig. \ref{fig:qual}, we compare the predictions of a generic and fixed joint-trained model (Sec. \ref{subsec:jt}) with our Highlight-TTA built on JAV \cite{badamdorj2021joint} on a test video from YouTube and TVSum. Highlight-TTA improves predictions over joint-trained model and is better aligned with the ground-truth.

% -------------------  Qualitative FIGURE
% ------------

\subsection{Ablation Studies}

To analyze the impact of each component of our method, Highlight-TTA, we conduct extensive ablation studies and compare them to several baseline methods. For our ablation studies, we experiment with our Highlight-TTA framework built upon the highlight detection network, JAV \cite{badamdorj2021joint}.

{\flushleft \textbf{Impact of meta-auxiliary learning and test-time adaptation (TTA):}} To investigate the effectiveness of meta-auxiliary learning (Meta-Aux) and test-time adaptation (TTA) in our framework, Highlight-TTA, we compare it with baselines where meta-auxiliary learning or TTA is not used. In Table \ref{table:ablation_meta}, we first report the results of our joint-trained model (Joint), which is jointly trained on both the primary highlight detection loss and the auxiliary cross-modality hallucinations loss. Next, we examine the effect of using the cross-modality hallucinations auxiliary task for updating the joint-trained model during test time directly, without the meta-auxiliary training step. It is noteworthy that this simple TTA method using cross-modality hallucinations already outperforms existing methods, demonstrating the effectiveness of using cross-modality hallucinations as an auxiliary task for highlight detection. Finally, in the last row, we present the results from our proposed method, Highlight-TTA, which integrates joint pre-training using both primary and auxiliary losses, meta-auxiliary training, and test-time adaptation. Here, the joint-trained model first undergoes meta-auxiliary training to produce a more adaptive and flexible model for test-time adaptation, allowing it to effectively fine-tune its parameters based on the internal information of each test video instance.

%  -------------- Ablation Meta
\setlength{\tabcolsep}{2pt}
\begin{table}[!h]
\centering
% \small
\footnotesize
% \begin{tabular}{@{}lcccc@{}}
    \begin{tabular}{ccccccc}
    \toprule
    \multicolumn{3}{c}{Method} & TVSum   & YouTube  & \multicolumn{2}{c}{QVHighlights \textit{val}} \\ 
    \cmidrule(lr){1-3} \cmidrule(lr){4-4} \cmidrule(lr){5-5} \cmidrule(lr){6-7}
    Joint & Meta-Aux & TTA   & top-5 mAP  & mAP             & mAP        & HIT@1     \\ \midrule
    \cmark & & & 68.30    & 71.20    & 24.12    & 33.16  \\
    \cmark &  & \cmark & 69.12 & 71.74 & 24.13  & 33.10 \\
    \cmark & \cmark &  & 70.25 & 72.03 & 24.30  & 33.10 \\
    \cmark & \cmark & \cmark & \textbf{71.05} & \textbf{73.02} & \textbf{24.31}  & \textbf{33.35} \\ 
    \bottomrule
    \end{tabular}
\caption{Ablation on the impact of meta-auxiliary learning (Meta-Aux) and test-time adaptation (TTA).}
\label{table:ablation_meta}
\end{table}
%----------------------------

\noindent{\textbf{Number of gradient updates:}} In Table \ref{table:ablation_num_updates}, we examine the impact of the number of gradient updates using the cross-modality hallucinations auxiliary task in the inner loop of meta-auxiliary training (Algo. \ref{alg:training}) and test-time adaptation of Highlight-TTA (Sec. \ref{subsec:tta}). We use the same number of gradient updates in both training and testing, which is intuitive and has been found useful in prior work \cite{chi2021test,hatem2023test}. Overall, our model performs best with three gradient updates, allowing the model to sufficiently adapt to the internal information of each test instance. Further increasing the number of updates does not yield any additional performance boost.

%  -------------- Ablation hal tta
\setlength{\tabcolsep}{2pt}
\begin{table}[htb]
\centering
% \small
\footnotesize
\begin{tabular}{ccccc}
% \begin{tabular}{C{1.75cm}*{4}{C{1.75cm}}}
\toprule
\multicolumn{1}{c}{\multirow{2}{*}{No. of gradient updates}} & TVSum           & YouTube         & \multicolumn{2}{c}{QVHighlights \textit{val}} \\ \cmidrule(lr){2-2} \cmidrule(lr){3-3} \cmidrule(lr){4-5}
\multicolumn{1}{c}{}                        & top-5 mAP       & mAP             & mAP        & HIT@1     \\ \midrule
1 & 68.24   & 71.59    & 24.26    & 31.74  \\
2  & 69.62 & 72.18 & 24.07 & 32.13 \\ 
3  & \textbf{71.05} & \textbf{73.02} & \textbf{24.31} & \textbf{33.35} \\ 

\bottomrule
\end{tabular}
\caption{Impact of number of gradient updates on model performance for meta-auxiliary training and TTA.}
\label{table:ablation_num_updates}
\end{table}
%  --------------

\noindent{\textbf{Performance on videos with missing audio modality:}} We conduct an experiment in which we randomly drop the audio modality for 25\% of the test videos. In this scenario, thanks to our auxiliary task of cross-modal hallucinations, we can substitute the missing audio information with hallucinated audio features using visual information. For a test video without audio modality, we discard the audio self-attention layer and replace the self-attended audio features in the proposed method with a detached copy of the output features of cross-modal audio hallucination module (see Figure \ref{fig:network}). Even in this challenging setting, Highlight-TTA remains helpful and outperforms the baseline. (Table \ref{table:ablation_missing_audio}).

% Ablation Missing Modality
%  -------------- Ablation missing _audio
\setlength{\tabcolsep}{2pt}
\begin{table}[htb]
\centering
% \small
\footnotesize
    \begin{tabular}{L{3.25cm}ccccc}
    \toprule
    \multicolumn{1}{l}{\multirow{2}{*}{Method}} & TVSum           & YouTube         & \multicolumn{2}{c}{QVHighlights \textit{val}}\\ \cmidrule(lr){2-2} \cmidrule(lr){3-3} \cmidrule(lr){4-5}
    \multicolumn{1}{c}{}                        & top-5 mAP       & mAP             & mAP        & HIT@1     \\ \midrule
    JAV & 68.42    & 70.18    & 23.99    & 32.32  \\
    \textbf{JAV + Ours} (25\% no audio) & \textbf{69.03}	& \textbf{71.25} & \textbf{24.26} &	\textbf{32.90} \\
    \bottomrule
    \end{tabular}
    \caption{Effect of missing audio modality in some test videos.}
    \label{table:ablation_missing_audio}
\end{table}
%  --------------

\noindent{\textbf{Analyzing robustness to corruption using Gaussian noise:}} To validate the robustness of our method under significant distribution shifts caused by noise during testing, we conduct the following experiment: we add randomly generated Gaussian noise to the input visual and audio features of each test video and evaluate the performance of our method. Despite the distribution shifts caused by the added noise, our TTA method, Highlight-TTA, still improves the performance of JAV \cite{badamdorj2021joint} across all three datasets (Table \ref{table:ablation_corruption}).

% Ablation Corruptions
%  -------------- Ablation corruption
\setlength{\tabcolsep}{1.2pt}
\begin{table}[htb]
\centering
% \small
\footnotesize
    \begin{tabular}{L{3.2cm}ccccc}
    \toprule
    \multicolumn{1}{l}{\multirow{2}{*}{Method}} & TVSum           & YouTube         & \multicolumn{2}{c}{QVHighlights \textit{val}}\\ \cmidrule(lr){2-2} \cmidrule(lr){3-3} \cmidrule(lr){4-5}
    \multicolumn{1}{c}{}                        & top-5 mAP       & mAP             & mAP        & HIT@1     \\ \midrule
    JAV & 68.42    & 70.18    & 23.99    & 32.32  \\
    \textbf{JAV + Ours} (with noise) & \textbf{69.21} &	\textbf{70.95}	& \textbf{24.21}	& \textbf{32.90}  \\
    \bottomrule
    \end{tabular}
    \caption{Impact of noise in test videos.}
    \label{table:ablation_corruption}
\end{table}
%  --------------

\noindent{\textbf{Impact of training data amount on meta-auxiliary training:}} To analyze the impact of the amount of training data, we randomly drop 25\% 
and 10\% of the training data during meta-auxiliary training. Despite having limited training data, Highlight-TTA remains effective and improves performance as more training data becomes available (Table \ref{table:ablation_limited}).

% Ablation limited data
%  -------------- Ablation limited
\setlength{\tabcolsep}{1.5pt}
\begin{table}[htb]
\centering
% \small
\footnotesize
    \begin{tabular}{L{3cm}ccccc}
    \toprule
    \multicolumn{1}{l}{\multirow{2}{*}{Method}} & TVSum           & YouTube         & \multicolumn{2}{c}{QVHighlights \textit{val}}\\ \cmidrule(lr){2-2} \cmidrule(lr){3-3} \cmidrule(lr){4-5}
    \multicolumn{1}{c}{}                        & top-5 mAP       & mAP             & mAP        & HIT@1     \\ \midrule
    JAV & 68.42    & 70.18    & 23.99    & 32.32  \\
        \textbf{JAV + Ours} (25\% train data dropped) & 69.10	& 71.36	& 24.01 & 32.32  \\
        \textbf{JAV + Ours} (10\% train data dropped) & 69.47 &	72.03 &	24.13	& 32.90 \\                             
    \textbf{JAV + Ours} & \textbf{71.05}  &	\textbf{73.02}  &	\textbf{24.31}  &	\textbf{33.35}  \\ \bottomrule
    \end{tabular}
    \caption{Impact of training data amount on meta-auxiliary training.}
    \label{table:ablation_limited}
\end{table}

%------------------------

\noindent{\textbf{Comparison with other TTA methods:}}
To evaluate the effectiveness of Highlight-TTA, in Table \ref{table:ablation_hal_tta}, we compare it with several popular TTA methods, including TENT \cite{wang2020tent}, EATA \cite{niu2022EATA}, and DeYO \cite{DeYOlee2024entropy}. TENT minimizes the entropy of the predictions at test time. EATA improves TENT by avoiding unreliable samples for adaptation. The more recent method, DeYO, proposes a new confidence metric and uses it along with entropy to dynamically re-weight each sample's contribution in TTA. Furthermore, we compare Highlight-TTA with a simple confidence-based pseudo-label technique \cite{lee2013pseudo,wang2022debiased,xie2020unsupervised,nassar2023protocon}, using this technique as an auxiliary task instead of the cross-modality hallucinations in our Highlight-TTA. Specifically, for each unlabeled test instance, our trained model iteratively generates pseudo-labels indicating highlight and non-highlight clips based on its high-confidence predictions, which are then used as targets to train the model. Highlight-TTA, which employs cross-modality hallucinations as an auxiliary task and meta-auxiliary training, achieves superior performance.

% Ablation TTA
%  -------------- Ablation hal tta
\setlength{\tabcolsep}{2pt}
\begin{table}[htb]
\centering
% \small
\footnotesize
    \begin{tabular}{lccccc}
    \toprule
    \multicolumn{1}{l}{\multirow{2}{*}{Method}} & TVSum           & YouTube         & \multicolumn{2}{c}{QVHighlights \textit{val}}\\ \cmidrule(lr){2-2} \cmidrule(lr){3-3} \cmidrule(lr){4-5}
    \multicolumn{1}{c}{}                        & top-5 mAP       & mAP             & mAP        & HIT@1     \\ \midrule
    JAV + Pseudo-label & 66.71    & 69.84    & 24.26    & 32.71  \\
    JAV + TENT \cite{wang2020tent} & 69.12    & 70.36    & 23.99    & 32.19  \\
    JAV + EATA \cite{niu2022EATA} & 69.81    & 70.67    & 24.09    & 32.77  \\
    JAV + DeYO \cite{DeYOlee2024entropy} & 67.87    & 70.96    & 24.03   &  32.52  \\
    \textbf{JAV + Ours} & \textbf{71.05} & \textbf{73.02} & \textbf{24.31}  & \textbf{33.35} \\ \bottomrule
    \end{tabular}
    \caption{Comparison with other TTA methods.}
    \label{table:ablation_hal_tta}
\end{table}
%  --------------

\noindent{{\textbf{Computational efficiency:}}} In Table \ref{table:computation_efficiency}, we compare the average inference time per video and GPU memory consumption of different TTA methods on YouTube. EATA and DeYO involve additional computations, such as calculating Fisher information matrices and performing object-destructive transformations, which result in higher memory usage. Highlight-TTA achieves state-of-the-art performance while maintaining a runtime and computational load that are comparable with the baseline TTA methods.
\begin{table}[h!]
\setlength{\tabcolsep}{8pt}
\centering
% \small
\footnotesize
    \begin{tabular}{L{2.3cm}C{1.5cm}C{2.8cm}}
    \toprule
    Method
    & Time (s) & GPU Memory (MB)      \\ 
    \midrule
    JAV + TENT \cite{wang2020tent} & 0.092 & 624    \\
     JAV + EATA \cite{niu2022EATA} & 0.082 & 646    \\
      JAV + DeYO \cite{DeYOlee2024entropy} & 0.089  & 642    \\
    \textbf{JAV + Ours} & 0.088 & 638	 \\ \bottomrule
    \end{tabular}
    \caption{Comparison of efficiency with other TTA methods.}
    \label{table:computation_efficiency}
\end{table}

\noindent{\textbf{TTA in cross-dataset settings:}}
To evaluate the adaptability of Highlight-TTA in a challenging scenario, we perform TTA in cross-dataset settings. Specifically, we use our meta-auxiliary trained JAV model on TVSum and evaluate it on YouTube, both directly (Meta-Aux) and with test-time adaptation (Meta-Aux + TTA). Similarly, we use the meta-auxiliary trained JAV model on YouTube and evaluate it on TVSum, both directly (Meta-Aux) and with TTA (Meta-Aux + TTA). The results in Table \ref{table:cross-dataset} demonstrate the effectiveness of Highlight-TTA in adapting to distribution shifts at test time and handling diverse videos.

\begin{table}[!htb]
    % \small
\footnotesize
    \begin{minipage}{.5\linewidth}
      \centering
        \begin{tabular}{l|c}
        \toprule
        \multicolumn{2}{c}{TVSum $\rightarrow$ YouTube}\\ 
        \midrule
        Meta-Aux & 62.59 \\
        Meta-Aux + TTA & \textbf{64.16}\\ 
        \bottomrule
        \end{tabular}
        
    \end{minipage}%
    \begin{minipage}{.5\linewidth}
      \centering
        \begin{tabular}{l|c}
        \toprule
        \multicolumn{2}{c}{YouTube $\rightarrow$ TVSum}\\ 
        \midrule
        Meta-Aux & 64.63 \\
        Meta-Aux + TTA & \textbf{66.44}\\ 
        \bottomrule
        \end{tabular}
    \end{minipage} 
    \caption{Impact of Highlight-TTA in cross-dataset settings.}
    \label{table:cross-dataset}
\end{table}

%------------------------

\noindent{\textbf{Is TTA useful for highlight detection?}} To assess its usefulness, we examine whether there is a distribution shift between training and test samples using the Frechet Inception Distance (FID) score \cite{heusel2017gans} for each dataset. FID compares the distribution of feature vectors extracted from real and generated images in GANs, with lower FID scores indicating greater similarity between the distributions of generated and real images. For our analysis, we replace the Inception features in the FID computation with the concatenated visual and audio features of each video, which are extracted using pre-trained models used in our method. We randomly split the training set into two parts, train (p1) and train (p2), and compute the FID score between these two parts. We then compute FID scores between the test set and train (p1), as well as between the test set and train (p2). The results reveal higher FID scores between the training parts and the test set compared to the FID score between the two training parts, indicating a greater distribution shift between the training and test sets of the datasets (Table \ref{table:ablation_distribution}). This underscores the importance of TTA in highlight detection, a key focus of our paper and the first effort in this area.

% ablation_distribution
\begin{table}[h!]
\setlength{\tabcolsep}{2pt}
\centering
% \small
\footnotesize
    \begin{tabular}{l|c|c|c}
    \toprule
    Dataset   & train (p1), train (p2)  & train (p1), test & train (p2), test \\
    \hline
    TVSum &	102.18 & \textbf{143.86} & \textbf{126.10} \\
    YouTube &	9.61  & \textbf{17.02}	  &   \textbf{15.79} \\
    QVHighlights	& 11.07 &	\textbf{14.13}	& \textbf{14.54} \\ \bottomrule
    \end{tabular}
    \caption{FID scores comparison between training and test split.}
    \label{table:ablation_distribution}
\end{table}
%------------------------

\section{Conclusion}\label{sec:conclude}
In this paper, we propose Highlight-TTA, a novel test-time adaptation (TTA) framework for video highlight detection that dynamically adapts the highlight detection model during testing to better align with the specific characteristics of each test video. Our approach leverages cross-modality hallucinations as an auxiliary task and incorporates meta-auxiliary learning, leading to performance gains for state-of-the-art highlight detection models across several benchmark datasets. The results underscore the potential of TTA in advancing highlight detection tasks and open up avenues for further research in this domain.

\noindent\textbf{Acknowledgements}: Zahidul Islam and Mrigank Rochan acknowledge the support of the University of Saskatchewan and the Natural Sciences and Engineering Research Council of Canada (NSERC).

{
    \small
    \bibliographystyle{ieeenat_fullname}
    \bibliography{highlight_refs_v2}
}

\end{document}